\documentclass[11pt, a4paper, onecolumn, draftcls]{IEEEtran}

\usepackage[utf8]{inputenc}
\usepackage{amssymb,amsmath,url}
\usepackage[linesnumbered,ruled,lined,boxed]{algorithm2e}
\usepackage[noend]{algpseudocode}

\usepackage[caption=false]{subfig}
\usepackage{graphicx}
\usepackage{xcolor}

\newcommand{\atcp}[1]{\tcp*[r]{\parbox[t]{.475\linewidth}{#1\hfill}}}

\SetCommentSty{mycommfont}

\begin{document}

\title{Least square ellipsoid fitting using iterative orthogonal transformations}

\author{
  \IEEEauthorblockN{Amit Reza, Anand S. Sengupta}\\
  \IEEEauthorblockA{Indian Institute of Technology Gandhinagar \\
                    Palaj Simkheda Gandhinagar Gujarat 382355, India.
                    }\\
  \{amit.reza, asengupta\}@iitgn.ac.in
}

\maketitle

\begin{abstract}
We describe a generalised method for ellipsoid fitting against a minimum set of data points. The proposed method is numerically stable and applies to a wide range of ellipsoidal shapes, including highly elongated and arbitrarily oriented ellipsoids. This new method also provides for the retrieval of rotational angle and length of semi-axes of the fitted ellipsoids accurately. We demonstrate the efficacy of this algorithm on simulated data sets and also indicate its potential use in gravitational wave data analysis.

\end{abstract}


\section{Introduction}

Reconstructing 3-dimensional (3D) ellipsoidal surfaces from discrete data points is a well-studied problem in the field of computer vision, pattern recognition, astronomy and medical image processing. Such 3D ellipsoid models find use in a broad class of applications: (a) One of the primary goals of computer vision is to find a suitable 3D shape descriptor for arbitrary shapes while preserving geometrical information as much as possible. Existing shape representations based on spherical descriptors are limited in scope as they are appropriate only for nearly spherical bodies. On the other hand, the ellipsoidal descriptors provide a closer approximation to irregular 3D shapes. Khatun et al. \cite{Asma} have proposed a method where ellipsoidal 3D shape representations have been used to retrieve arbitrary 3D shapes. (b) Gait analysis is utilised for a systematic study of animal locomotion by measuring and tracking their body mechanics. Gathering gait features and interpretation of the gait dynamics is a challenging task as current techniques are computationally expensive. 3D ellipsoid fitting methods have been found useful in this line of research. Sivapalan et al. \cite{Sivapalan} have proposed a fast 3D ellipsoid based gait recognition algorithm using a 3D voxel model, (c) galaxies are often modelled as 3-dimensional ellipsoids whose parameters are determined from images recorded by telescopes. Comp{\`e}re et al. \cite{Paul} have recently proposed a three-dimensional galaxy fitting algorithm to extract parameters of the bulge, long bar, disc and a central point source from broadband images of galaxies.

Several techniques exist for fitting ellipsoids to a set of data points and can be broadly classified into projection based algorithms \cite{Rita, Kayikcioglu} and nonlinear optimisation based surface fitting algorithms \cite{Keren, Taubin}. Projection based fitting algorithms are also further organised into two categories: namely, orthographic and line integral based. 

The basic idea in orthographic projection is to use matrix operators to project a 3D shape onto planes. For the case of a 3D ellipsoid, three different orthographic transformations are possible along three orthogonal planes, and projected shapes are 2D ellipses. If the parameters of the projected ellipses are deciphered, then 3D rotation between two successive projections can be detected after characterising the variations of the semi-axes length and the orientation of the projected ellipses. On the other hand, the line integral projection based methods commonly use projection contours to reconstruct ellipsoids. The general second-degree equation of an ellipsoid is used to construct the line integral projection model. 
One of the shortcomings of such methods is that one needs prior information about the projected ellipses to determine the angle and axis of rotation accurately.

In nonlinear optimisation techniques for fitting ellipsoidal surfaces, the latter is modelled as a bounded surface through a family of polynomials which are then fitted using standard nonlinear optimisation methods. The problem with such techniques is that due to high non-linearity of the model, the optimal solution may get stuck in local solutions leaving the resulting surface unbounded. Therefore the proper solution can not guarantee closed bounded solution of the desired surface. To overcome this problem Li et al. \cite{Li} have prescribed an algorithm to obtain closed form of the resulting ellipsoidal surface by providing an additional constraint. The fitting algorithm works robustly for ellipsoids whose short radius is at least half of their major radius and for which the semi-axes of the model ellipsoid are aligned along the co-ordinates. More recently, Ying et al. (2012) \cite{Ying} have proposed a least-square ellipsoid fitting algorithm by extending the 2-dimensional ellipse fitting algorithm given by Fitzgibbon et al. \cite{Fitzgibbon}. 

In this work, we propose a stable algorithm that can fit an ellipsoidal surface to a given set of data points and can detect the rotational angle as well as semi-axes length with significant improvement over Li (2004) \cite{Li}.  The new method is applicable even to extreme cases where the ellipsoid is highly elongated and arbitrarily oriented to a rigid frame of reference. Our primary motivation is to extend their idea in such an algorithmic form, which can produce the best fit for \textit{any kind} of the ellipsoidal surface. Also, we describe a method for the retrieval of the orientation of such ellipsoids without assuming any prior information.

 This paper is organised as follows: In Section-II, we present a concise description of Li et al. \cite{Li}, establishing the notation used in this article and highlighting the salient features of their algorithm.
 Section-III describes our proposed method based on the general equation of an ellipsoid.
 In section-IV, we describe the algorithm for retrieval of the orientation of the reconstructed ellipsoid, followed by a demonstration of the efficacy of our method using synthetic data. 
 We then present a case study where this approach is applied to the field of gravitational wave data analysis in Section-V.
 
 Finally, we make some general comments on the results obtained in this paper.

\section{Previous Work:}
\label{sec:leastSq}

The general equation of the second degree in three variables $(x,y,z)$ representing a conic is given by:

 \begin{equation}\label{SecDeg}
  ax^2+by^2+cz^2+2fyz+2gxz+2hxy+2px+2qy+2rz+d = 0.
 \end{equation}

As shown in \cite{Li}, Eq. (\ref{SecDeg}) represents an ellipsoid under the constraint 
\begin{equation}
kJ-I^2 = 1 .
\label{Const}
\end{equation}
where
\begin{equation}\label{Cond1}
 I \equiv a+b+c,
\end{equation}
\begin{equation}\label{Cond2}
 J \equiv ab+bc+ac-f^2-g^2-h^2,
\end{equation}
and $k$ is a positive number. For ellipsoids with comparable semi-axes lengths, $k \sim 4$.

Let P =  $\mathbf{p_{i}(x_{i},y_{i},z_{i})},\{$i = 1,2,..,N$\}$ be the coordinates of $N$ points with respect to a fixed frame of reference $XYZ$ (refer Figure \ref{fig:gtf_right}) to which an ellipsoid is to be fitted. Further, let the ellipsoid be arbitrarily oriented in this frame.
For every point $\mathbf p_{i}$, one defines a column $\mathbf{X_{i}}$ of the {\em design matrix} $\mathbf D$ as:
\begin{equation}\label{Coord}
  \mathbf{X_{i}} = (x_{i}^2,y_{i}^2,z_{i}^2,2y_{i}z_{i},2x_{i}z_{i},2x_{i}y_{i},2x_{i},2y_{i},2z_{i},1)^T.
\end{equation}
To fit an ellipsoidal surface, each data point must satisfy the quadratic Eq. (\ref{SecDeg}) with the constraint 
defined in Eq. (\ref{Const}). The algebraic distance $\Omega$ between the model and the set of data points defined as,
 \begin{equation}\label{AlgDis}
  \Omega = \sum_{i = 1}^N(\mathbf{v}^T\mathbf{X_i})^2
 \end{equation}
must be minimized with respect to $\mathbf v$ in order to find the best fit, where
 \begin{equation}\label{Param}
  \mathbf{v} \equiv (a,b,c,f,g,h,p,q,r,d)^T 
 \end{equation} is the set of unknown parameters whose values are to be determined.
Therefore the ellipsoid fitting problem can be mapped to an optimisation problem that can be solved using standard least square methods.

It is obvious that Eq. (\ref{SecDeg}) can be written in {\it matrix form} as the following system of 
linear equations: 
\begin{equation}\label{LnEq}
   \mathbf{D}^T \mathbf{v} = 0,
\end{equation}
in terms of the design matrix $\mathbf{D} = (X_{1},X_{2},...,X_{i})$ of order $10 \times N$, where $N \geq 10$. 
The geometric distance above can also be written in matrix form as $\Omega = || {\mathbf {D v}} ||^2 = \mathbf{v}^T\mathbf{D}^T\mathbf{vD}$, 
which is to be minimized subject to the constraint given in Eq. (\ref{Const}). The latter can also be written in matrix form as
$\mathbf{v}^T\mathbf{Cv} = 1$, where 
\begin{equation}\label{Contour}
\mathbf{C} =
  \begin{bmatrix}
    -1 & \frac{k}{2}-1 & \frac{k}{2}-1 & 0 & 0 & 0 & 0 & 0 & 0 & 0\\
    \frac{k}{2}-1 & -1 & \frac{k}{2}-1 & 0 & 0 & 0 & 0 & 0 & 0 & 0\\
    \frac{k}{2}-1 & \frac{k}{2}-1 & -1 & 0 & 0 & 0 & 0 & 0 & 0 & 0\\
    0 & 0 & 0 & -k & 0 & 0 & 0 & 0 & 0 & 0\\                                 .
    0 & 0 & 0 & 0 & -k & 0 & 0 & 0 & 0 & 0\\
    0 & 0 & 0 & 0 & 0 & -k & 0 & 0 & 0 & 0
  \end{bmatrix}.
\end{equation}

The Lagrangian of this optimization problem is defined as,
\begin{equation}\label{ConstMin}
\begin{aligned}
& & \mathcal{L}{(\mathbf{v},\lambda)} = \Omega-\lambda (\mathbf{v}^T\mathbf{Cv}-1),\\
\end{aligned}
\end{equation}
where $\lambda$ is the scalar Lagrange multiplier.
Using the standard Lagrange multiplier method \cite{Strang}, we set ${\partial \mathcal{L}}/{\partial \bf v} = 0$ and ${\partial \mathcal{L}}/{\partial \lambda} = 0$ leading to
\begin{equation}\label{GnEig}
 \mathbf{DD}^T\mathbf{v} = \lambda \mathbf{Cv},
\end{equation}
and
\begin{equation}\label{ConstMat}
 \mathbf{v}^T\mathbf{Cv} = 1,
\end{equation}
respectively.
 
Eq. (\ref{GnEig}) is in the form of a generalized eigenvalue equation which can be solved for $\lambda$ and $\bf{v}$.
The eigenvectors and corresponding eigenvalues can be used to determine the semi-axes length and its orientation as explained later in this paper. 
The optimal value of the parameter $k$ in Eq. (\ref{Const}) depends on the input data set: for a given data set; the optimisation leads to the correct value of $k$ which is to be determined iteratively.

Ying et al. (2012) \cite{Ying} have also established a similar type of generalised eigenvalue equation by introducing the concept of a random plane that intersects the quadratic surface defined in Eq. (\ref{SecDeg}). Under this paradigm, a quadratic surface is deemed to be an ellipsoid if its intersection with any random plane is an ellipse. Incorporating this idea leads to some modifications to the constraint matrix for the generalised eigenvalue system in Eq. (\ref{GnEig}). The eigenvector corresponding to a unique positive eigenvalue provides the required solution.

\section{Methodology} 
\label{sec:methodology}

Eq. (\ref{SecDeg}) can be normalized by $d(\neq 0)$. Therefore the actual number of unknown parameters involved in the system is 9. Further, if the centre of the ellipsoid is known, one can fix it at the origin $(0,0,0)$ without any loss of generality, in which case, the number of unknown parameters further reduces to 6. Thus a minimum of six unique data points is sufficient to find $\mathbf{v}$ unambiguously.

Eq. (\ref{SecDeg}) can be written in matrix form as:
\begin{equation}\label{Contour1}
 \mathbf{A^{T}KA} = -d
\end{equation}
where 
\begin{equation}\label{CovMat}
\mathbf{K}=
  \begin{bmatrix}
    a & h & g &  \\
    h & b & f &  \\
    g & f & c                
  \end{bmatrix},
\end{equation}

\begin{equation}
 \mathbf{A} = 
    \begin{bmatrix}
    x & y & z
   \end{bmatrix}.
\end{equation}

The Fisher information matrix $\mathbf{K}$ is constructed from the elements of $\mathbf{v}$, and our aim in fitting the ellipsoid is to reconstruct this matrix robustly from the given data points. The eigenvectors of $\bf{K}$ are aligned along the principal directions of the ellipsoid. Off-diagonal terms signify cross-correlation
among the variables and allude to the fact that these axes are not aligned along the rigid frame of reference $XYZ$.
Conversely, $f = g = h = 0$ implies that principle axes of the ellipsoid are aligned along the rigid frame of reference.

Starting from an initial estimate, we aim to find a conformal transformation through a rotation matrix $\mathbf R$ in an iterative fashion, in which $\bf K$ becomes diagonal. Note that rotation matrices must be orthogonal, i.e. $\mathbf R \mathbf{R^T} = \mathbf I$.

An initial estimate of $\mathbf K$ can be made by using the fact that the Fisher information matrix is equal to the inverse of the data covariance matrix. Uniform sampling of data points over the ellipsoid can lead to a good initial estimate of $\bf K$ by this method, but pathological cases may arise when all the data points are sampled from a narrow region on the ellipsoidal surface. In the latter case, the inverse of the data covariance matrix (if it exists) may not be a good initial estimate of $\bf{K}$. The corresponding $\mathbf R$ is constructed from the eigenvectors of $\mathbf K$.

Alternatively, $\mathbf R$ can be constructed from a $\it{random}$ positive definite matrix $\mathbf{G} \in \mathcal{N}(0,1)$ in such a way that each column of $\mathbf{R}$ is formed from the eigenvectors of the covariance of $\mathbf{G}$. 

\begin{figure}%
    \centering   
    
    \subfloat[Absolute error in sexi-axes calculation in each iteration.]{\label{fig:Err_Axes}%
                {\includegraphics[width=0.75\textwidth]{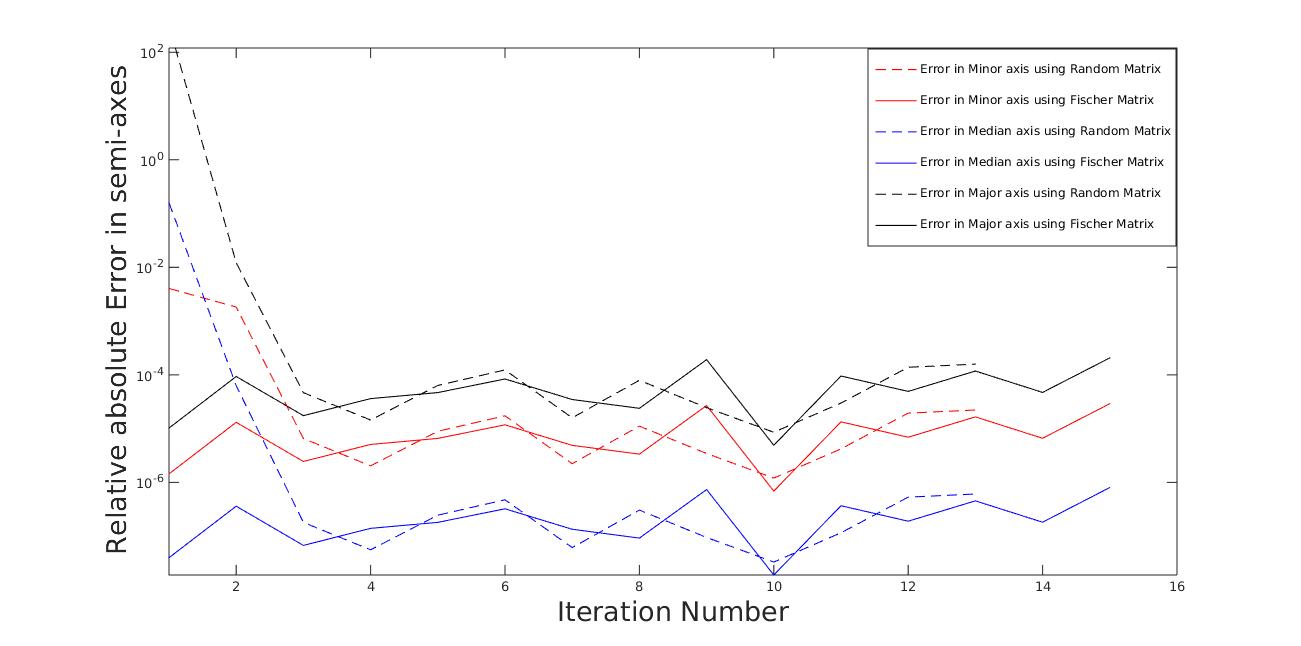} }}%
    \quad   
    \subfloat[Absolute error in angles calculation in each iteration.]{\label{fig:Err_Ang}%
                {\includegraphics[width=0.75\textwidth]{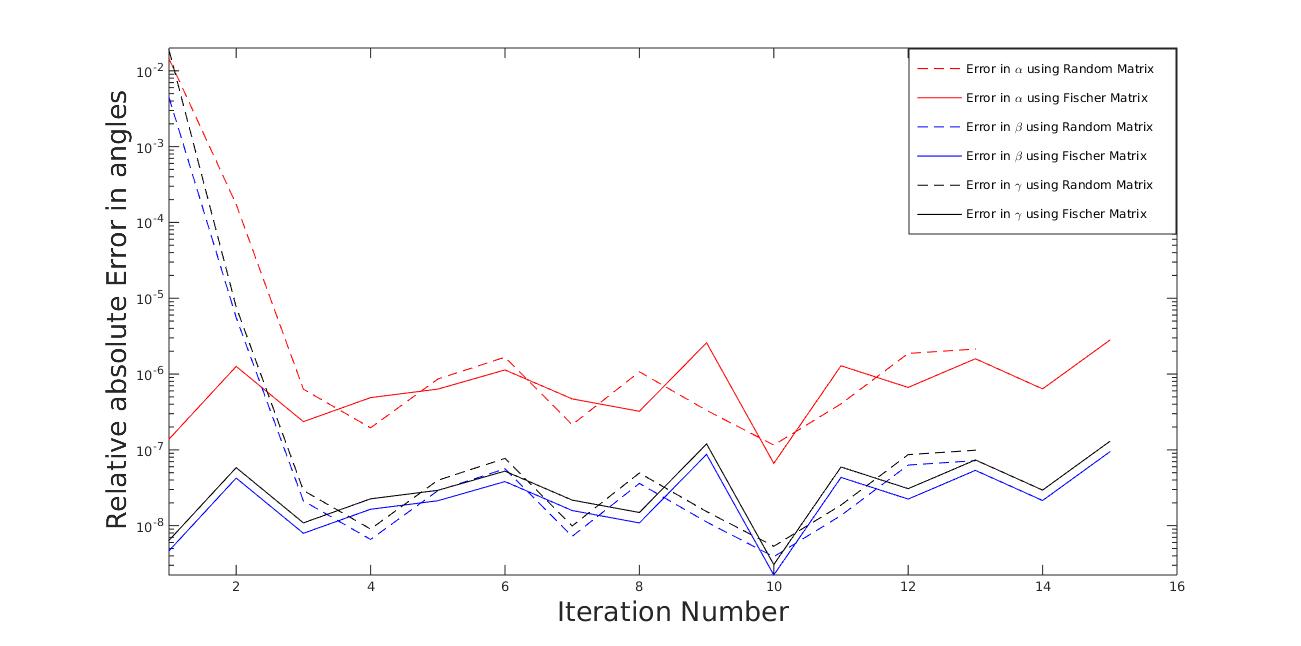}}}%
                                          
    \caption{
    Convergence of the iterative transformation algorithm to reconstruct ellipsoids given minimal (six) data sets. In this case, the input values
    ($A$, $B$, $C$, $\alpha$, $\beta$, $\gamma$) correspond to extremely high values of $\chi$ ($10^4$). 
    }
    \label{fig:err}%
\end{figure}

 Regardless of the method used to initialize $\mathbf R$, we show in Fig \ref{fig:Err_Axes}-\ref{fig:Err_Ang} that our algorithm achieves the convergence criteria leading to the diagonal form of $\bf{K}$ after a few iterative steps.
Since no prior information is assumed about the way in which data points are sampled from the surface, it is advisable to initialise $\bf R$ using the random matrix method.

In a single iteration of this algorithm, $\mathbf R$ is used to project the data points which are then used to find the best fitting ellipsoid using the method of least squares as outlined in Section \ref{sec:leastSq}. This process leads to the best fit $\mathbf K$ whose eigenvectors are used to further refine $\mathbf R$. This process continues until the desired termination criteria are met. In the process, the fixed axes $XYZ$ undergoes a series of successive conformal transformations until it aligns with the principal directions of the ellipsoid. The successive updates to 
$\mathbf{R}$ are recorded and used to reconstruct $\mathbf K$ in $XYZ$ by applying an inverse transformation: $\mathbf K_{XYZ} = \mathbf{R^T} \mathbf{K R}$. 

\begin{figure}%
    \centering   
    
    \subfloat[Rotated ellipsoid]{\label{fig:gtf_left}%
                {\includegraphics[width=0.4\textwidth]{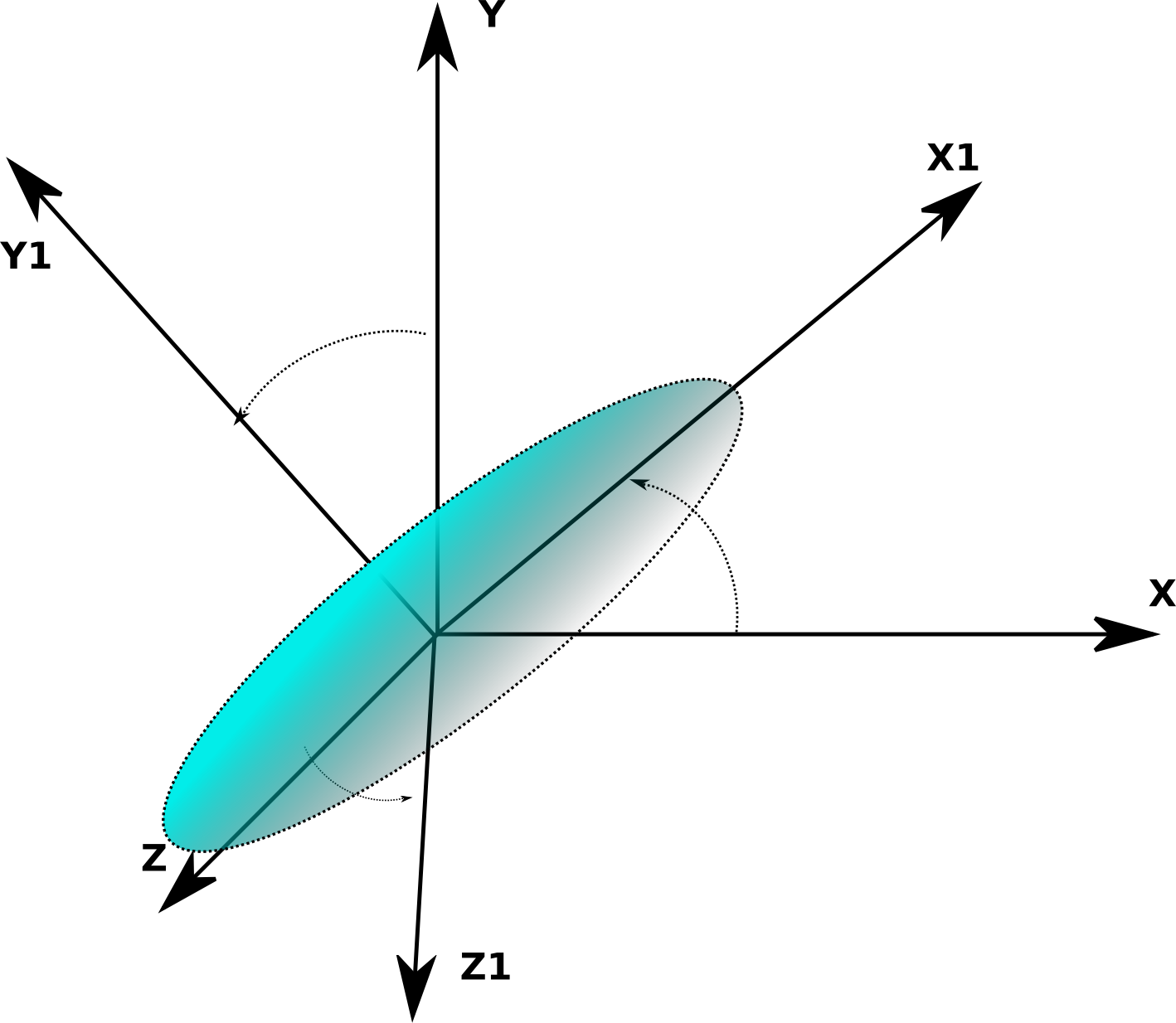} }}%
    \quad   
    \subfloat[Estimated principal axes in each iterative stage.]{\label{fig:gtf_right}%
                {\includegraphics[width=0.4\textwidth]{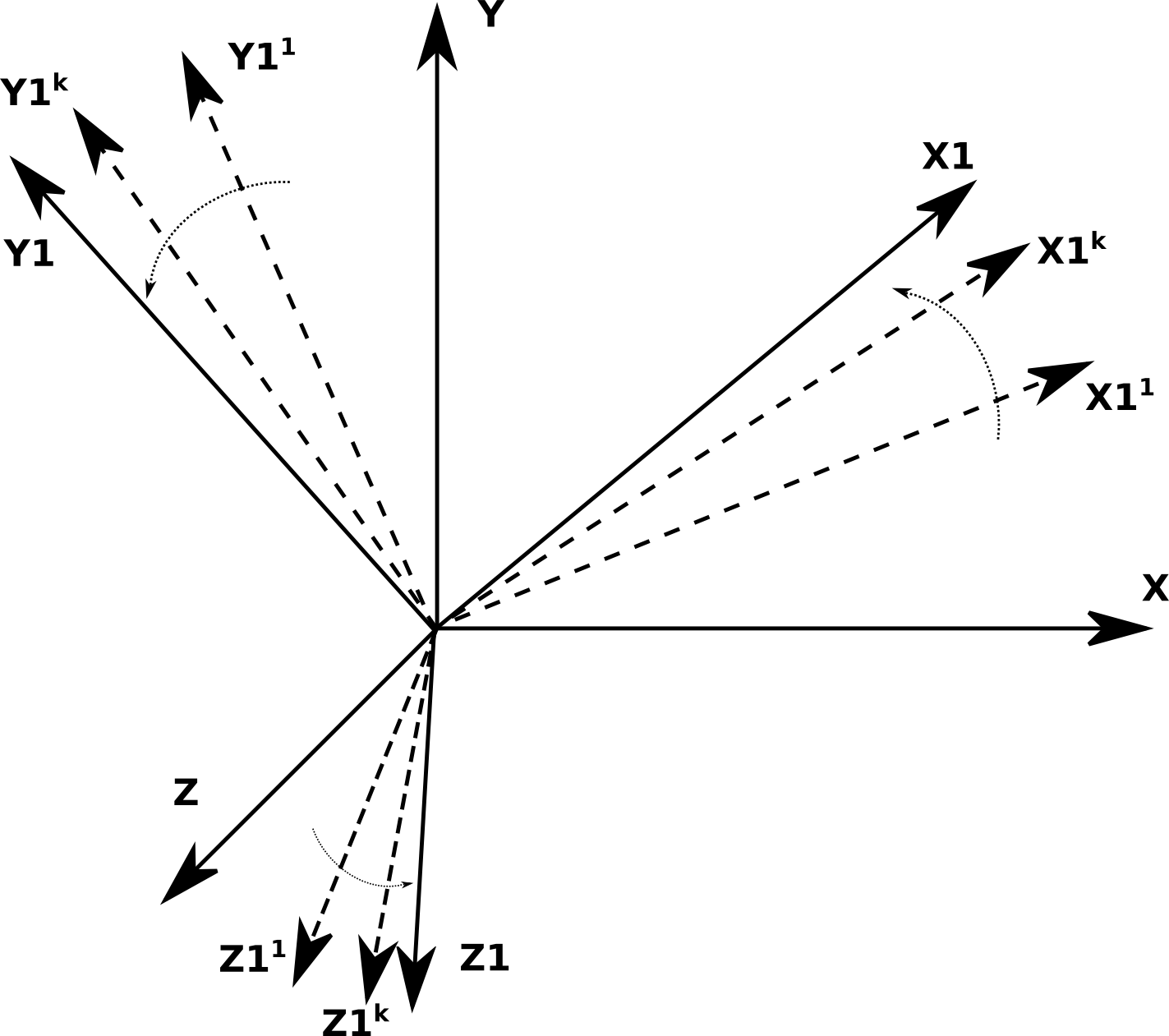}}}%
                                          
    \caption{
      Schematics of the iterative conformal transformations: $XYZ$ represents the initial frame of reference w.r.to which the coordinates of the data points are supplied, and $X_1Y_1Z_1$ are actual principal axes of the ellipsoid. $X_1^kY_1^kZ_1^k$ represents the estimated fixed frame of reference after $k^{th}$ iterative stage to which the data points are projected.}
    \label{fig:gtf}%
\end{figure}

The steps of this method are given in Algorithm \ref{alg:main} and illustrated in Figure \ref{fig:gtf}. 

\subsection{Convergence criteria}
\label{sec:Convergence Criteria}

As stated above, the iterative algorithm proceeds by projecting the data points through a succession of conformal transformations until $\mathbf K$ becomes diagonal. At convergence, the off-diagonal terms $f, g, h$ become nearly zero, and the trace of the eigenvector matrix\footnote{Eigen vector matrix refers to the matrix whose columns are the eigenvectors in a decreasing order of the eigenvalues} of $\mathbf K$ is nearly equal to $3$.

The initial estimate of $\mathbf R$ may be quite off the mark in which case; the least square fitting algorithm may require an enormous value of $k$ in the constraint equation Eq. (\ref{Const}) for meeting the least-square convergence criteria. It can even lead to a failure to find the optimal least square solution. This can be solved by restricting $k$: $4\leq k\leq k_{max}$ to a maximum upper limit $k_{max} \sim 10^{10}$. As a consequence, the least square ellipsoid fittings in the early stages of the iterative algorithm may be sub-optimal, but it does not affect the outcome. As successive conformal transformations are applied, one requires progressively smaller values of $k$ for convergence at every iteration. Restricting $k$ to an upper limit also makes it computationally more efficient. 

If the ellipsoid is aligned along fixed frame, then an initial choice of $\bf R = \mathcal{I}_{3 \times 3}$ will further reduce the number of iterative transformations to a one step fitting problem.

\subsection{Efficient retrieval of orientation angles}

A general Euler rotation matrix is of the following form.
\begin{equation}\label{rot1}
\mathbf{R} = 
  \begin{bmatrix}
    R_{11} & R_{12} & R_{13}\\
    R_{21} & R_{22} & R_{23}\\
    R_{31} & R_{32} & R_{33}\\
  \end{bmatrix}
\end{equation}
This matrix can be factorized as a product of rotations in the following sequence:
\begin{equation}\label{rot2}
    \mathbf{R} = \mathbf{R_{Z}}(\gamma)\mathbf{R_{Y}}(\beta)\mathbf{R_{X}}(\alpha),
\end{equation}
where $\alpha$, $\beta$, $\gamma$ represents the Euler angles corresponding to $X$, $Y$, and $Z$ axes respectively.
The elements $R_{ij}$ of $\mathbf R$ represent a specific function of the Euler angles as given in Eq. (\ref{rot3}). 
These functions can be inverted to solve for $\alpha$, $\beta$, $\gamma$. Note that these are an overdetermined set of 9 equation in 3 unknowns. We may use any three to solve for the Euler angles and use the others for consistency check.

As we do not assume any prior information about the orientation and shape of the ellipsoid, we initialize $\mathbf R$ randomly which undergoes a series of refinements through the iterative fitting procedure. At the end of this fit, the fixed frame $XYZ$ when transformed through $\mathbf R$ aligns itself along the ellipsoid axes. However, this alignment can happen in 6 different ways. This is easily understood from Figure \ref{fig:gtf_left} - the transformed X-axis can align along $\pm X_1$ or $\pm Y_1$ or $\pm Z_1$ directions. The convention used for determining the Euler angles \cite{Slabaugh} is such that the transformed $X$ axis lies along the principal $X_1$ direction of the ellipsoid. Therefore, as highlighted in Algorithm  \ref{proc:chal1}, one must scan through all 6 column permutations of the rotation matrix $\mathbf R$ to re-orientate the axes correctly.

\begin{algorithm}[H]
\DontPrintSemicolon

\KwIn{$\mathbf {\{p_i\}} \in \mathbf{R}_{{N\times 3}}: N \geq 6 $ \atcp{\parbox[t]{.5\linewidth}{Data points}} }

$\mathbf{R}\gets \mathbf{G}$ \atcp{Initialize $\mathbf R: \mathbf R \mathbf {R^T} = \mathbf I$,  using a positive definite matrix $\mathbf G  \in \mathcal{N}$(0,1)}

set $k_{max} = 10^{10}$\;

\While {True} {
    $\mathbf {\{p_i\}} = \mathbf{\{p_i\}}\times \mathbf R $ \atcp{Transformation of the data points}
  
    $\mathbf{D} \gets \mathbf{X_i} \gets \mathbf{\{p_i\}}$   \atcp{Design matrix}    
  
    Set $k = 4$
    
    \While {True} {
    
      $\mathbf{K} \gets  \mathrm{Solve}\; \{\mathbf{DD}^{T} = \lambda \mathbf{C v} \}$\;
    
      \uIf{$(\mathrm{lsqConv} \footnote{lsqConv is the converge criteria for least-square ellipsoid fit for given k as given in Eq. (\ref{Const})}\; ||\;  k > k_{max})$} { 
      
        break                        
      }
      \Else{
        $k \gets 2 \times k$
      }
    }
  
    $\mathbf{R} = \mathbf{R} \times \mathbf{[evec[\mathbf K]]}$  \atcp{Refine $\mathbf{R}$} 
  
    \If{$(\mathrm{conv.} \footnote{The conv. criteria tested refers to those in Section-IIIA})$} {
    
      break 
    }       
}

$ \mathbf{K} = \mathbf{R} \times (\mathbf{K}/{d}) \times \mathbf{R}^{T}$  \atcp{Inverse transform to the frame XYZ}

$(\hat{A},\hat{B},\hat{C}) = 1/\sqrt{\lambda_{i}}$, $\lambda_{i} \gets eig[\bf{K}]$  \atcp{Estimate of semi-axes}      

\textbf{Return:} {$\hat{A}$, $\hat{B}$, $\hat{C}$, $\mathbf R$, $\mathbf K$}\;

\caption{(Semi-axes, $\mathbf{R}$, $\mathbf{K}$) = \textbf{FitEllipsoid}($\mathbf {\{p_i\}}$): Iterative transformation based least square ellipsoid fitting}
\label{alg:main}
\end{algorithm}

\begin{algorithm}[H]

\SetKw{Return}{return}
\Indp
\DontPrintSemicolon 
\KwIn{$\{\mathbf {p_{i}}\}$, $\mathbf R$. }

\For {$n = 1$ $\mathrm{:}$ $6$} {
    $\mathbf{\tilde{R}}$ = $\mathrm{Reshuffle}$\footnote{$\mathrm{Reshuffle}$ ($\mathbf{R}$,n) returns the $n^{th}$ parmutation of the column vectors of $\mathbf{R}$.} ($\mathbf R, n $)
    
    $(\tilde \alpha, \tilde \beta, \tilde \gamma)$ $\gets$ $\mathbf{\tilde{R}}$
    
    $\mathbf{\tilde  R'} = \mathbf{\tilde  R_z} (\tilde \gamma) \mathbf{\tilde  R_y} (\tilde \beta) \mathbf{\tilde  R_x} (\tilde \alpha) $
    
    $\mathbf{\tilde{\{p_{i}\}}} \gets $ $\mathrm{Reshuffle}$($\mathbf{\{p_{i}\}}, n$)$\times \mathbf{\tilde{R}'}$
    
    \If {$\mathbf{\tilde{\{p_{i}\}}}$ = $\mathbf {\{p_{i}\}} $} {
      \Return $(\tilde \alpha, \tilde \beta, \tilde \gamma)$
    } 
}
\Indm
\caption{Sub-routine to calculate Euerler angles from given rotation matrix $\mathbf R$ and data $\{p_{i}\}$. }
\label{proc:chal1}
\end{algorithm}

\subsection{Computational complexity}
\label{sec:complexity}

The proposed iterative method depends on two different convergence criteria (lsqConv (Eq. (\ref{Const})) and conv (\ref{sec:Convergence Criteria})). In the inner loop, the convergence criteria lsqConv optimises the value of $k$. The time complexity of this circuit depends on solving the generalised eigenvalue problem as given by Eq. (\ref{GnEig}). On the other hand, the conv criteria in the outer loop optimise the alignment of the principal axes of the fitted ellipsoid.

From Algorithm \ref{alg:main}, the total time complexity for each iteration in the inner loop can be seen to be $\mathcal{O}(10\; N^2 + 10^3)$, where $N$ is the number of supplied data points. The number of iterations depends on $\chi$. In our numerical experiments, we have noted that extremely elongated ellipsoids (large $\chi$) can take 35 – 40 iterations (with six input data points). On the other hand for smaller values of $\chi$, only a few iterations are enough. 

It is also evident that every iteration of the outer loop requires $\mathcal{O}(3N^2 + 3^3)$ operations from the matrix multiplications in Step $4$ and $15$ respectively. Just like the inner loop, the number of iterations for the outer loop also depend on $\chi$. In our experience with simulated data sets, this can be as high as 15 – 20 for extremely elongated ellipsoids. For input data from moderately elongated ellipsoids, the outer loop is seen to converge within 2 – 4 iterations. 

The proposed algorithm is robust even for a small number of data points: results presented in this paper are for the minimum set of data points.

\section{Implementation and Experimental Results}

In this section, we describe the generation of synthetic data points to confront it against the new algorithm and present the results. We also compare these results against Li's (2004) \cite{Li} and Ying's (2012) \cite{Ying} method wherever possible.

\subsection*{Synthetic Data set}

Let $\alpha$, $\beta$, $\gamma$ be the Euler angles and $A$, $B$, $C$ the predefined semiaxes length of the ellipsoid. The data points $\{p_i\}, \ i=(1,2,\ldots,N\geq 6)$ on the surface of this ellipsoid are generated most conveniently in polar coordinates: 
\begin{equation}\label{eq:data}
\{ p_i \} = {\mathbf{R}}^{T} \;
  \begin{bmatrix}
    A\cos \theta_{i} \cos \phi_{i} & B\cos \theta_{i} \sin \phi_{i} & C\sin \theta_{i}\\
  \end{bmatrix}
\end{equation}
where the angles $\theta_{i}$ and $\phi_{i}$ are generated from uniformly distributed random numbers in the interval $[0,\pi]$ and $[0,2\pi]$ respectively. 
$\bf R$ is the rotation matrix whose elements are given in terms of the Euler angles as:
\begin{equation}\label{rot3}
\begin{split}
    R_{11} &= \cos \alpha \cos \beta, \\
    R_{12} &= \sin \gamma \sin \beta \cos \alpha - \cos \gamma \sin\alpha, \\
    R_{13} &=  \cos \gamma \sin \beta \cos \alpha + \sin \gamma \sin \alpha,\\
    R_{21} &= \cos \beta \sin \alpha, \\
    R_{22} &= \sin \gamma \sin \alpha \sin \beta + \cos \gamma \cos\alpha, \\ 
    R_{23} &= \cos \gamma \sin \alpha \sin \beta - \sin \gamma \cos \alpha, \\
    R_{31} &= -\sin \beta, \\
    R_{32} &= \sin \gamma \cos \beta, \\
    R_{33} &= \cos \gamma \cos \beta.\\ 
\end{split}   
\end{equation}

We generate several sets of data for aligned as well as arbitrarily oriented ellipsoids corresponding to different ellipsoidal shapes characterised by
a parameter $\chi = A/C$; defined as the ratio of major and minor axes. Different data sets were generated corresponding to $\chi$ values ranging from $\chi \sim 1.5 $ to $\chi \sim 10^4$ to test this algorithm. A particular case of interest are extremely elongated
and flat ellipsoids with a very high value of $\chi$ - such ellipsoids arise in the context of gravitational wave data analysis and are separately discussed in the next section.

\subsection*{Results}

Some sample results of the ellipsoid reconstruction tests using synthetic data sets are summarised in the tables below. 

The first column of each table are input semi-axes length and Euler angles $A,B,C,\alpha,\beta,\gamma$ used to generate the data points using Eq (\ref{eq:data}).
The second and third columns are the estimated values of these quantities using the algorithm described in this paper. As mentioned earlier, there are two independent ways to make the initial guess for $\mathbf R$ used to project the data. One involves initial estimation from the eigenvectors of the Fisher information matrix (calculated from input data) whereas the other involves initializing $\mathbf R$ using a random positive definite matrix whose elements are drawn from $\mathcal{N}(0,1)$. The tables show the result of both these cases. The fourth column contains the estimate of ellipsoid parameters using Li's algorithm and serves as a baseline. Data sets 1-3 tabulate sample results where the input data is generated from an ellipsoidal surface whose principal axes are aligned with $XYZ$. Data sets 4-6  are sample results for the non-aligned case.

For aligned cases corresponding to small values of $\chi$ (e.g. Data Set-1), both algorithms (the one presented in this paper and Li's (2004)) were able to reconstruct the ellipsoid accurately. But for $\chi$ values $\geq 5$ (e.g. Data Sets-2, 3, 5, 6), Li's method was observed to reconstruct the ellipsoid incorrectly in certain cases. This issue becomes more pronounced as we increase $\chi$. For non-aligned ellipsoids with $\chi \sim 10^2$ or more, Li's method gave different answers for every new set of random input points. On the other hand, the algorithm presented here was robust even for extreme ellipsoidal shapes arbitrarily oriented to the fixed axes.

The algorithm as given by Ying X et al. \cite{Ying} works robustly for ellipsoidal surfaces whose principal axes are aligned along the fixed frame of reference. It is seen to perform well for extremely elongated and flat ellipsoids (large values of $\chi$). But for arbitrarily oriented ellipsoids, it was observed that the method was not able to reconstruct the ellipsoids correctly. Therefore the method is very sensitive and numerically unstable for non-aligned ellipsoidal surfaces.

\begin{table}[ht]
\caption{Data Set 1 ($\chi=1.5$, aligned)} 
\centering
\begin{tabular}{|c|c||c|c|c|l}

\cline{1-5}
\multicolumn{1}{ |c| }{Input} & \multicolumn{2}{ c| }{This work} & Li(2004) & Ying(2012) \\ \cline{2-3}

& Fisher matrix & Random matrix& & \\ \cline{1-5}

\multicolumn{1}{ |c|  }{$A = 12.0$ } & $12.0$ & $12.0$  & $12.0$  & $12.0$            \\ \cline{1-5}

\multicolumn{1}{ |c| }{ $B = 10.0$} & $10.0$ & $10.0$  & $10.0$   & $10.0$            \\ \cline{1-5}

\multicolumn{1}{ |c| }{ $C = 8.0$} & $8.0$ & $8.0$ & $8.0$        & $8.0$              \\ \cline{1-5}

\multicolumn{1}{ |c|  }{$\alpha = 0.0, \beta = 0.0, \gamma = 0.0$} & $\alpha = 0.0, \beta = 0.0, \gamma = 0.0$ & $\alpha = 0.0, \beta = 0.0, \gamma = 0.0$  & -  & -        \\ \cline{1-5}

\end{tabular}
\end{table}

\begin{table}[ht]
\caption{Data Set 1 ($\chi = 5$, aligned)} 
\centering
\begin{tabular}{|c|c||c|c|c|l}

\cline{1-5}
\multicolumn{1}{ |c| }{Input} & \multicolumn{2}{ c| }{This work} & Li(2004) & Ying(2012)  \\ \cline{2-3}

& Fisher matrix & Random matrix& & \\ \cline{1-5}

\multicolumn{1}{ |c|  }{$A = 5.0$ } &$5.0$ & $5.0$  & $4.086188652$ & $5.0$             \\ \cline{1-5}

\multicolumn{1}{ |c| }{ $B = 3.0$} & $3.0$ & $3.0$  & $3.586034787$ & $3.0$              \\ \cline{1-5}

\multicolumn{1}{ |c| }{ $C = 1.0$} & $1.0$ & $1.0$ & $2.684837382$  & $1.0$               \\ \cline{1-5}

\multicolumn{1}{ |c|  }{$\alpha = 0.0, \beta = 0.0, \gamma = 0.0$} & $\alpha = 0.0, \beta = 0.0, \gamma = 0.0$ & $\alpha = 0.0, \beta = 0.0, \gamma = 0.0$  & -   & -       \\ \cline{1-5}
\end{tabular}
\end{table}

\begin{table}[ht]
\caption{Data Set 3 ($\chi = 10$, aligned)} 
\centering
 \begin{tabular}{|c|c||c|c|c|l}
\cline{1-5}
\multicolumn{1}{ |c| }{Input} & \multicolumn{2}{ c| }{This work} & Li(2004) & Ying(2012)   \\ \cline{2-3}

& Fischer matrix & Random matrix& &  \\ \cline{1-5}

\multicolumn{1}{ |c|  }{$A = 10.0$ } & $10.0$ & $10.0$ & $10.050239317$ &$10.0$               \\ \cline{1-5}

\multicolumn{1}{ |c| }{ $B = 6.0$} & $6.0$ & $ 6.0$  & $5.103102823$    &$6.0$                \\ \cline{1-5}

\multicolumn{1}{ |c| }{ $C = 1.0$} & $1.0$ & $1.0$ & $4.547883756$      &$1.0$                 \\ \cline{1-5}

\multicolumn{1}{ |c|  }{$\alpha = 0.0, \beta = 0.0, \gamma = 0.0$} & $\alpha = 0.0, \beta = 0.0, \gamma = 0.0$ & $\alpha = 0.0, \beta = 0.0, \gamma = 0.0$  & - & -        \\ \cline{1-5}

\end{tabular}
\end{table}

\begin{table}[ht]
\caption{Data Set 4 ($\chi=1.5$, non-aligned)} 
\centering
\begin{tabular}{|c|c||c|c|c|l}

\cline{1-5}
\multicolumn{1}{ |c| }{Input} & \multicolumn{2}{ c| }{This work} & Li(2004) & Ying (2012)  \\ \cline{2-3}

& Fisher matrix & Random matrix&  &\\ \cline{1-5}

\multicolumn{1}{ |c|  }{$A = 12.0$ } & $12.0$ & $12.0$  & $12.0$  &  $16.4749$          \\ \cline{1-5}

\multicolumn{1}{ |c| }{ $B = 10.0$} & $10.0$ & $10.0$  & $10.0$   &  $9.2354$           \\ \cline{1-5}

\multicolumn{1}{ |c| }{ $C = 8.0$} & $8.0$ & $8.0$ & $8.0$        &  $7.6336$           \\ \cline{1-5}

\multicolumn{1}{ |c|  }{$\alpha = 30$} & $30.0$ &$30.0$& -        & -          \\ \cline{1-5}

\multicolumn{1}{ |c| }{$\beta = 80$ } & $80.0$  &$80.0$& -        & -       \\ \cline{1-5}

\multicolumn{1}{ |c| }{$\gamma = 70$} & $70.0$  &$70.0$& -        & -          \\ \cline{1-5}

\end{tabular}
\end{table}
\begin{table}[ht]
\caption{Data Set 5 ($\chi= 5$, non-aligned)} 
\centering
\begin{tabular}{|c|c||c|c|c|l}

\cline{1-5}
\multicolumn{1}{ |c| }{Input} & \multicolumn{2}{ c| }{This work} & Li(2004) & Ying(2012)  \\ \cline{2-3}

& Fisher matrix & Random matrix & & \\ \cline{1-5}

\multicolumn{1}{ |c|  }{$A = 1.0$ } & $0.999257113$ & $ 0.999128747$  & $1.280262088$ &  $0.9129$           \\ \cline{1-5}

\multicolumn{1}{ |c| }{ $B = 3.0$}  & $3.002419519$ & $3.002838810$   & $2.550536762$ &  $1.2791$          \\ \cline{1-5}

\multicolumn{1}{ |c| }{ $C = 5.0$}  & $5.000012418$ & $5.000014725$   & $5.058022788$ &  $1.3335$         \\ \cline{1-5}

\multicolumn{1}{ |c|  }{$\alpha = 70$} & $69.980565$ &$69.977193$  & -  & -         \\ \cline{1-5}

\multicolumn{1}{ |c| }{$\beta = 10$ } & $9.989666$ &$9.987882$& -       & -      \\ \cline{1-5}

\multicolumn{1}{ |c| }{$\gamma = 30$} & $30.007390$ &$30.008666$& -     & -       \\ \cline{1-5}

\end{tabular}
\end{table}

\begin{table}[ht]
\caption{Data Set 6 ($\chi=10$, non-aligned)} 
\centering
\begin{tabular}{|c|c||c|c|c|l}

\cline{1-5}
\multicolumn{1}{ |c| }{Input} & \multicolumn{2}{ c| }{Our Method} & Li(2004) & Ying(2012)  \\ \cline{2-3}

& Fisher matrix & Random matrix&  & \\ \cline{1-5}

\multicolumn{1}{ |c|  }{$A = 10.0$ } & $10.0$ & $10.0$  & $5.076706206$  & $1.9486$           \\ \cline{1-5}

\multicolumn{1}{ |c| }{ $B = 3.0$} & $3.0$  & $3.0$  & $2.862531749$     & $1.5969$            \\ \cline{1-5}

\multicolumn{1}{ |c| }{ $C = 1.0$} & $1.0$ & $1.0$ & $2.492054846$       & $0.8945$            \\ \cline{1-5}

\multicolumn{1}{ |c|  }{$\alpha = 50$} & $50.0$ & $50.0$  & -            & -               \\ \cline{1-5}

\multicolumn{1}{ |c| }{$\beta = 60$ } & $60.0$ &$60.0$& -                & -                  \\ \cline{1-5}

\multicolumn{1}{ |c| }{$\gamma = 40$} & $40.0$ &$40.0$& -                & -                  \\ \cline{1-5}

\end{tabular}
\end{table}

\section{Case study: Application to Gravitational Wave Data Analysis}
\label{sec:gw}

 Gravitational wave data analysis aims to search for weak gravitational wave signals from compact massive astrophysical objects such as binary neutron stars and black holes. The direct detection of astrophysical gravitational wave signals has opened a new observational window to the physical Universe, 
 and is expected to complement the information obtained from traditional astronomy in various electromagnetic observation bands.
 
 The two advanced LIGO \cite{AdLIGO} gravitational-wave observatories made the first direct detection of these signals about a year ago on 14 September 2015 \cite{Abbott}.  Several other observatories are also being commissioned including LIGO-India \cite{ligo-india}. A network of three or more of such detectors is expected to improve the scientific potential of these searches by a large margin.
 
 Accurate gravitational wave signal from inspiral, merger and coalescence of compact binaries can be calculated \cite{Faye} theoretically. These theoretical models allow the well-known technique of matched filtering to be used for detecting faint signals buried in detector noise.  If the latter is Gaussian, then it can be shown that the matched filter is optimum, yielding maximum signal to noise ratio (SNR).
 
 The matched filtering strategy for gravitational wave searches is to compute the cross-correlation between the interferometer output and a  set of template waveforms; over the detector bandwidth, weighted inversely by the noise power spectrum of the detector  (\cite{Owen, Dhurandhar, Satya}). The deemed parameter space is gridded appropriately for adequate coverage, and the template waveforms are constructed for every point in this grid. The grid is also known as the bank of templates in gravitational wave literature. The construction of this template bank (in other words, the grid over the deemed parameter space) is aided by inducing a metric on the signal manifold. 
 
The match, or overlap between two templates $h(\vec \lambda_1)$ and $h(\vec \lambda_2)$ is defined through their inner-product integral
\begin{equation}\label{eq:match}
\left \langle h(\vec \lambda_1), \; h(\vec \lambda_2) \right \rangle  \equiv 4 \mathrm{Re}  \int_{0}^{\infty} df \; 
\frac{{\tilde{h}}^* (f; \vec{\lambda}_1) \; {\tilde{h}} (f; \vec{\lambda}_2)}{S_h(f)} ,
\end{equation}

where, $\sim$ over the symbols denotes frequency domain representation and $*$ denotes complex conjugation. The one-sided noise power spectral density is given by $S_h(f)$.

For two nearby templates $h(\vec{\lambda})$ and $h(\vec{\lambda} + \Delta {\vec \lambda})$ one can expand the match $M$ as a Taylor series around $\vec{\lambda}$, to get the following expression upto lowest order term in $\Delta {\vec \lambda}$:
 \begin{equation}\label{MT}
  M(\vec{\lambda},\Delta{\vec \lambda}) \approx 1 + \sum_{i, j=1}^N \frac{1}{2} \left ( \frac{\partial ^2 M}{\partial \Delta {\lambda^{i}} \partial \Delta {\lambda^{j}}} \right ) \biggr\rvert_{\Delta \vec \lambda = 0}\Delta {\lambda}^{i} \Delta {\lambda}^{j}. 
 \end{equation}

In the above expression, $\Delta \lambda^{i,j} \in \Delta \vec\lambda$. Further, we have normalized the templates such that $M(\vec{\lambda}, \Delta {\vec \lambda}=0) = 1$.
From Eq. (\ref{MT}) one can see that the $\it {mismatch}$ $(1-M)$ can be used to define the distance square 
$(\Delta s^2)$ between two nearby templates in terms of a metric $g_{ij}$ induced on the signal manifold as:
\begin{eqnarray}
\Delta s^2      &=& (1-M) , \\
              &=& \sum_{i, j=1}^N g_{ij}\Delta {\lambda^{i}} \Delta {\lambda^{j}},
              \label{eq:hyperellipsoid}
      \end{eqnarray}
where,
\begin{equation}\label{Metric}
  g_{ij} = -\frac{1}{2} \left (\frac{\partial ^2 M}{\partial\Delta {\lambda^{i}} \Delta {\lambda}^{j}} \right ) \biggr\rvert_{\Delta {\vec \lambda} = 0}.
 \end{equation}
It is clear that at a fixed minimal match, the above equation describes the surface of a hyper-ellipsoid (fixed centre).

For a $N$ dimensional signal manifold, $g_{ij}$ is a square symmetric matrix with $N(N+1)/2$ independent components. We pause to note that $g_{ij}$ is not constant over the signal manifold due to to the curvature of the space. This metric is widely used in the gravitational wave signal analysis for placement of templates, determination of consistency of triggers from multiple detectors \cite{Anand}, etc.

Our aim in this section is to demonstrate the efficient 
numerical estimation of $g_{ij}$ using the technique developed in earlier sections of this paper for the particular case of $N=3$.
The latter corresponds to the case where the signal is described by three parameters: two component masses and an effective mass weighted spin magnitude parameter \cite{Ajith} of the compact binary system. Conventionally, one reparameterizes these to new chirp time coordinates $(\tau_0, \tau_{3}, \tau_{3s})$ in which the metric is almost flat (slowly changing).
Comparing Eq. (\ref{eq:hyperellipsoid}) with Eq. (\ref{Contour1}), we immediately notice the correspondence between the metric $g_{ij}$ and the Fisher information matrix $\mathbf{K}$. 
In this case $g_{ij}$ at a fixed point $\vec \lambda$ has six independent components. One starts by numerically solving the match equation $\left \langle h(\vec{\lambda}), \; h(\vec{\lambda} + \Delta{{\vec \lambda}}) \right \rangle = M $ for $\Delta \vec \lambda$ along (at least) 6 random directions. 
These six points can now be used to fit the ellipsoidal surface Eq. (\ref{eq:hyperellipsoid}) using the iterative technique developed earlier. The Fisher information matrix corresponding to the ``best-fit'' ellipsoid gives the best numerical estimate of the metric $g_{ij}$ at the point $\vec{\lambda}$ in the parameter space. 

\begin{figure}
        \centering
        \includegraphics[width=.60\linewidth]{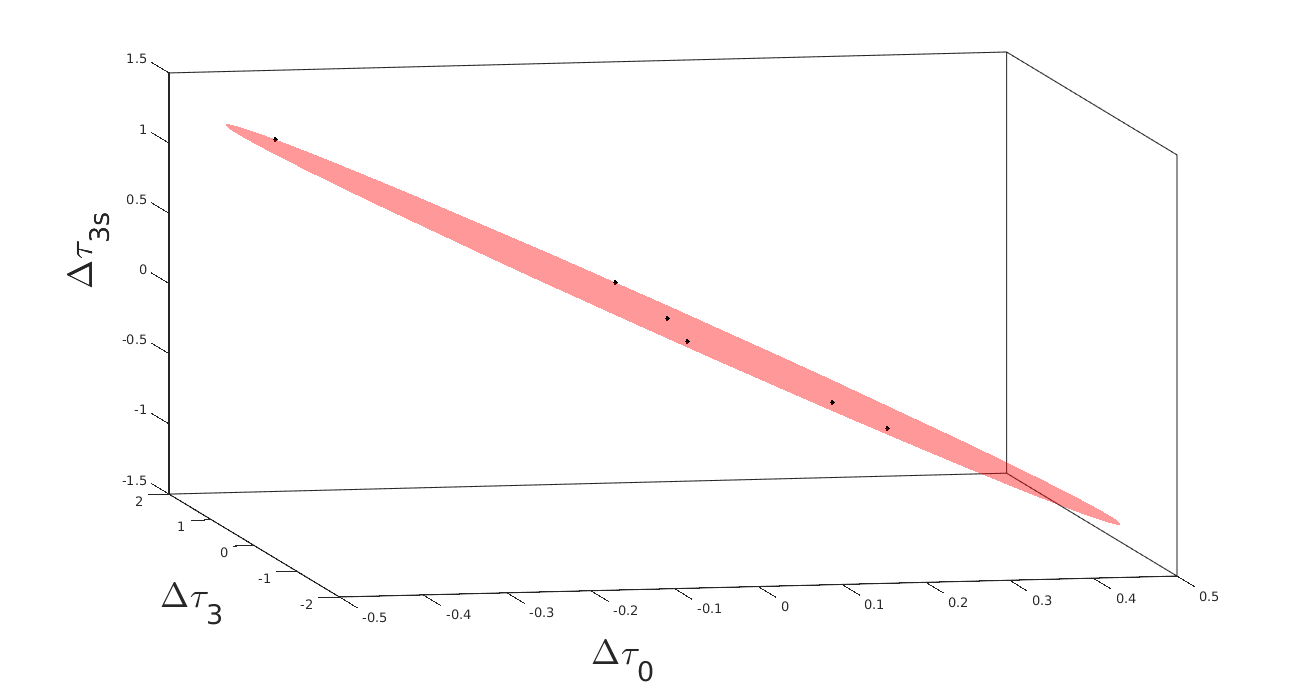}
        \caption{{Fitted constant match ellipsoid Eq. (\ref{eq:hyperellipsoid}) for TaylorF2RedSpin model of gravitational waveform from inspiraling compact binaries.}}        
\label{fig:TaylorF2RedSpin_Ellipsoid}
\end{figure} 

We demonstrate the ellipsoid fitting and numerical estimation of the metric $g_{ij}$ on the gravitational wave signal manifold,
at a point $\vec{\lambda}_0$ correspond to the component masses $(8.0,2.0)M_\odot$ and reduced spin magnitude $0.0$. The minimal match is taken to be $0.97$ and the advanced LIGO 'aLIGOZeroDetHighPower' model for $S_h(f)$ is assumed.
The match equation is solved along six randomly chosen directions from $\vec{\lambda}_0$ and used as test data points to fit the ellipsoid which is shown in Fig \ref{fig:TaylorF2RedSpin_Ellipsoid}. Table-\ref{tab:fitValues} shows the estimated semi-axes and Euler angles (in degrees) obtained from $g_{ij}$:

\begin{table}
    
    \caption{Reconstructed semi-axes and Euler angles for constant match ellipsoid as described in Section \ref{sec:gw}} 
    \centering
    \begin{tabular}{|r|l|l|c|c|c|}
    \hline
    $A = 2.14$  & $B = 0.047$  & $C = 0.004$ & $\alpha = 33.72$ & $\beta = 7.72$ & $\gamma = 19.53$ \\ \hline
    \end{tabular}
\label{tab:fitValues}
\end{table}
    
We notice that the ellipsoid has high value of $\chi \simeq \frac{2.14}{.004} = 535$.
The elements of the Fisher information matrix can also be estimated using semi-analytic techniques, using the classic formula 
\begin{equation}
\label{eq:semiAnalytic}
g_{ab} = \left \langle {\partial h}/{\partial {\lambda}_a}, \; {\partial h}/{\partial {\lambda}_b} \right \rangle; a,b= 1,2,3; \; \lambda_{a,b} \in \vec \lambda.
\end{equation}

\begin{figure}
       
        \centering
        \includegraphics[width=.75\linewidth]{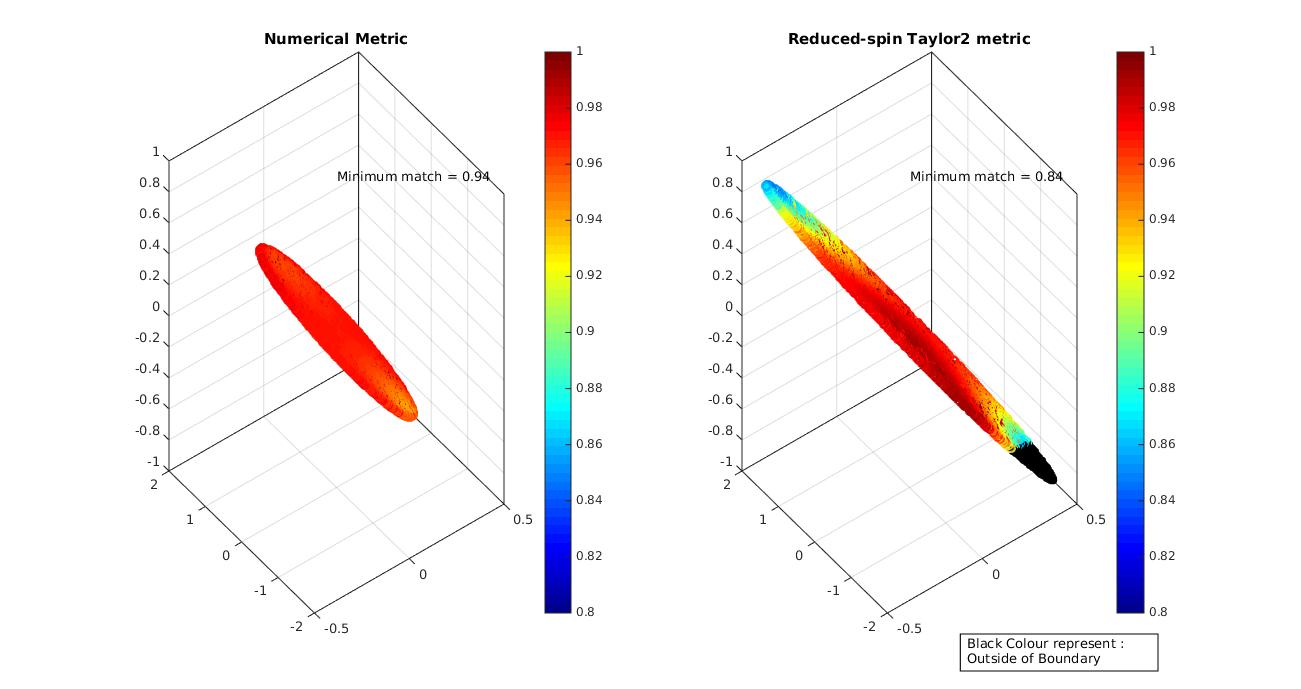}
        \caption{{Comparison of the constant match ($M=0.97$) ellipsoids generated semi-analytically (right panel) and numerically using the new method presented in this paper (left panel). The TaylorF2RedSpin model is assumed. The centre point is chosen to be ($80.0, 3.5, 0.0$) in the chirp time co-ordinates. $40,000$ points at random on both these ellipsoidal surfaces are chosen, and their match with the centre point is plotted. The colour bar in the figure indicates the numerical range of matches. This clearly shows that the numerically estimated metric is better as the match values are closer to the desired constant level.}}        
 \label{fig:ellipsoidCompare}
\end{figure}

We find that the numerical approach via ellipsoid fitting provides a better estimate of the metric as determined by the following test. The metric components are calculated at a fiducial point ($80.0, 3.5, 0.0$) in the chirp time ($\tau_0,\tau_3,\tau_{3s}$) coordinate space at a constant minimal match of $M=0.97$ in two different ways: (a) by semi-analytical technique outlined in Eq. (\ref{eq:semiAnalytic}) and (b) numerically, via the ellipsoid fitting algorithm presented in this work.  If the metric coefficients are correctly computed, then we expect the squared distance between a random point on the ellipsoidal surface and the centre to be $\leq (1-M)$. In other words, an arbitrary point on the ellipsoids should have a match $\geq M$ with the centre as evident from Eq. (\ref{Metric}). To this end, we sprayed $40,000$ points at random on both these ellipsoidal surfaces and plotted their match with the centre in Fig. \ref{fig:ellipsoidCompare}. The colour bar in the figure indicates the numerical range of matches. It is clear that the metric estimated by ellipsoidal fitting (left panel) is better in comparison to the semi-analytical method, as the matches are closer to the desired level (0.97). On the other hand, many parts of the semi-analytically obtained ellipsoid show match values that are significantly below this level (indicating over-estimation of the ellipsoid size). 

\section{Conclusion}
In this paper, we have developed a general algorithm for fitting ellipsoids of arbitrary shape and orientation using an iterative random transformation based method. We have shown the new method can fit long, thin or compressed ellipsoid and also able to retrieve the rotation angle accurately. Our method is based on iteratively improving the fit by changing the orientation of the coordinates to align along the axes of the ellipsoid. We have verified the accuracy and numerical stability of our algorithm using several sets of synthetic data.
Finally, we have demonstrated how this algorithm can be used to numerically estimate the metric on the signal manifold of gravitational wave signals.

\section*{Acknowledgements}

Amit Reza would like to thank Indian Institute of Technology Gandhinagar for a research fellowship. Thanks are also due to fellow PhD students Soumen Roy, Chakresh Kr. Singh and Md. Yousuf Jamal for useful discussions and help with the manuscript.

\bibliography{references}{}
\bibliographystyle{ieeetr}

\end{document}